\documentclass[sigconf,nonacm,balance=false]{acmart}

\usepackage{microtype}
\usepackage{graphicx}
\usepackage{subcaption}
\usepackage{booktabs}
\usepackage{multirow}
\usepackage{mathtools}
\usepackage[section]{placeins}
\usepackage{xcolor}
\usepackage{mdframed}
\usepackage{fvextra}

\graphicspath{{./images/}}
\setcopyright{none}
\renewcommand\footnotetextcopyrightpermission[1]{}
\renewcommand{\shortauthors}{Wang et al.}
\newmdenv[
  linecolor=gray!55,
  backgroundcolor=gray!8,
  linewidth=0.6pt,
  roundcorner=3pt,
  innertopmargin=8pt,
  innerbottommargin=8pt,
  innerleftmargin=10pt,
  innerrightmargin=10pt,
  skipabove=8pt,
  skipbelow=8pt
]{promptbox}

\fvset{
  breaklines=true,
  breakanywhere=true,
  fontsize=\small
}

\begin{document}
\raggedbottom

\title{LaP-Forensics: Latent-Pixel Consistency Guided Multimodal Reasoning for Deepfake Detection}

\author{Can Wang}
\affiliation{%
  \institution{The Hong Kong Polytechnic University}
  \city{Hong Kong}
  \country{China}
}

\author{Yuhao Wang}
\affiliation{%
  \institution{University College London}
  \city{London}
  \country{United Kingdom}
}

\author{Yushe Cao}
\affiliation{%
  \institution{Tsinghua University}
  \city{Beijing}
  \country{China}
}

\author{Canran Xiao}
\affiliation{%
  \institution{Sun Yat-sen University}
  \city{Guangzhou}
  \country{China}
}

\author{Fei Shen}
\authornote{Corresponding author.}
\affiliation{%
  \institution{National University of Singapore}
  \city{Singapore}
  \country{Singapore}
}

\begin{abstract}
Recent generative models can produce images with few obvious visual artifacts, weakening detectors and explanations that rely only on surface appearance. We present LaP-Forensics, a multimodal framework that augments RGB semantics with reconstruction-based forensic evidence. A frozen Stable Diffusion DDIM inversion--reconstruction model provides a fixed reconstruction reference, and its residual map measures local compatibility with that reference. Independent projectors encode the RGB image and residual map before a structured \textit{Where--What--Why} model predicts a textual analysis and an artifact mask. Supervised fine-tuning is followed by Group Relative Policy Optimization (GRPO), whose reward combines mask overlap with output-structure and evidence-reference terms. These text-side terms encourage the model to refer to the consistency map but do not constitute a verifier of free-form textual truth. A separate image-level head fuses RGB and DDIM-residual class features. Experiments show cross-generator detection on UniversalFakeDetect and competitive artifact localization on the official SynthScars benchmark. Controlled cue-construction, inversion-horizon, component, reward-term, and counterfactual analyses support the utility of the residual stream under the evaluated settings, while free-form textual faithfulness and reliability under post-processing remain open limitations. Source code is available at \url{https://github.com/Healgney/LaP-Forensics}.
\end{abstract}


%% ACM CCS Concepts -- required for final submission; not needed for review
%% Generate at: https://dl.acm.org/ccs

\renewcommand{\authors}{Can Wang\and Yuhao Wang\and Yushe Cao\and Canran Xiao\and Fei Shen}
\hypersetup{pdfauthor={Can Wang, Yuhao Wang, Yushe Cao, Canran Xiao, Fei Shen}}
\maketitle

\section{Introduction}
Recent advances in generative AI~\cite{shen2024advancing,shen2024imagpose}, exemplified by models such as Flux and Stable Diffusion 3.5, have substantially reduced the conspicuous artifacts that once made synthetic images easy to identify. As generated content becomes increasingly coherent at both semantic and local visual levels, human inspection and detectors relying on a narrow set of appearance cues struggle to provide reliable and generalizable forensic judgments~\cite{yan2025sanity}.

\begin{figure}[!t]
  \centering
  \includegraphics[width=\columnwidth]{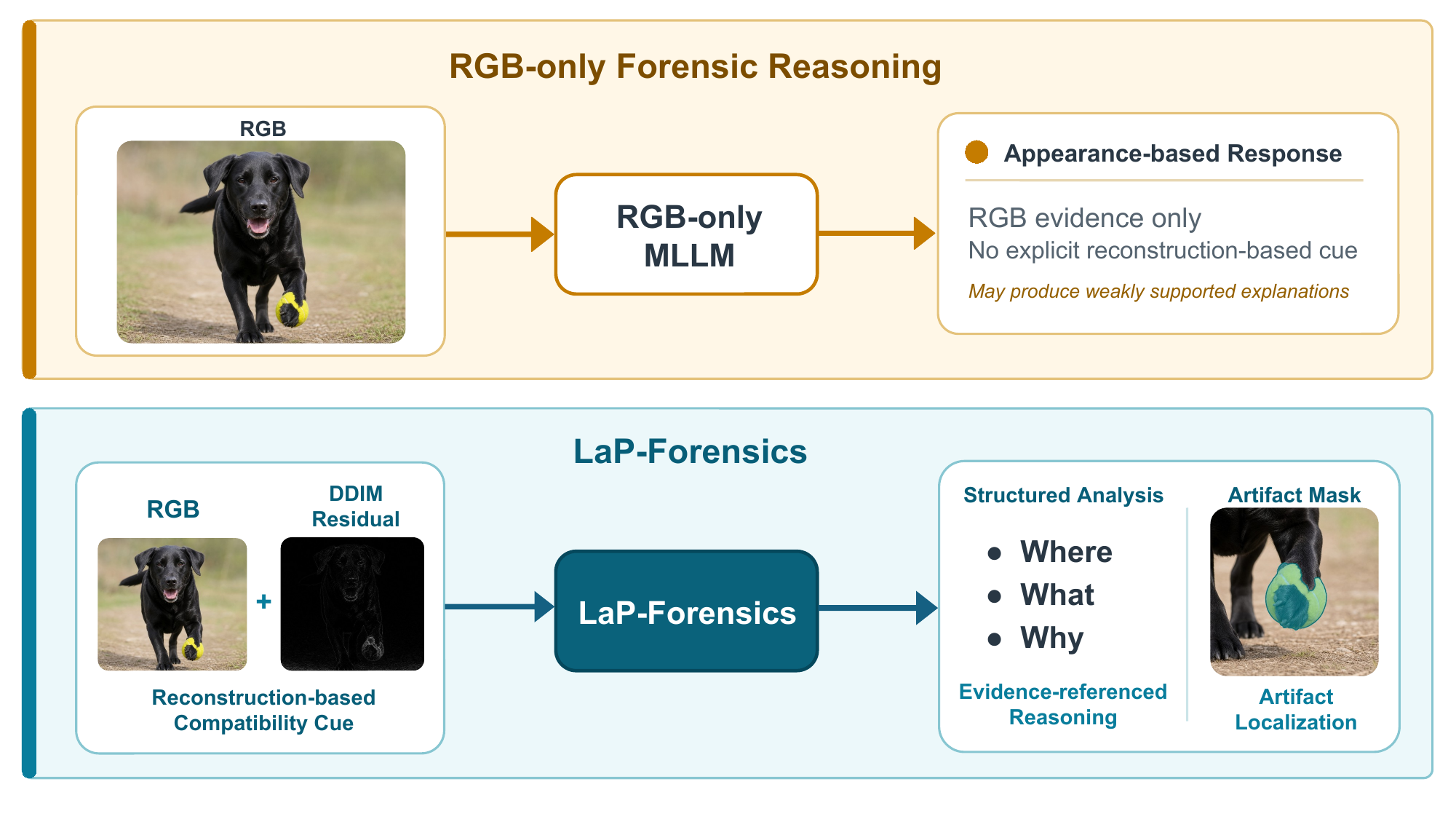}
\caption{Motivation for LaP-Forensics. RGB-only forensic reasoning relies primarily on appearance cues and lacks an explicit reconstruction-based reference. LaP-Forensics complements RGB semantics with a DDIM-derived compatibility residual computed against a fixed reconstruction reference, providing auxiliary forensic evidence for structured, evidence-referenced reasoning and artifact localization.}
  \label{fig:motivation}
\end{figure}

Recent advances in controllable image synthesis, including pose-guided generation, virtual dressing, fine-grained garment synthesis, and story visualization~\cite{shen2024advancing,shen2024imagpose,shen2025imagdressing,shen2025imaggarment,shen2025boosting}, have further increased the realism and diversity of synthetic visual content. Existing forensic approaches can be broadly divided into image-level detectors, such as FakeInversion~\cite{cazenavette2024fakeinversion}, and MLLM-based forensic agents, such as LEGION~\cite{kang2025legion} and ManipShield~\cite{xu2025manipshield}. Although image-level detectors can provide authenticity predictions, they typically offer limited semantic interpretation. In contrast, MLLM-based agents generate human-readable explanations, but their rationales may not be adequately grounded in explicit forensic evidence. As illustrated in Figure~\ref{fig:motivation}, an RGB-only model may attribute a prediction to an apparently malformed object even when the claimed artifact is absent or unsupported by the underlying forensic signal. This mismatch between the predicted label and the stated evidence undermines the reliability of free-form forensic explanations~\cite{ji2025interpretable,sun2025visuallinguistic}.
This limitation highlights the need for forensic models that not only predict image authenticity but also ground their explanations in explicit and spatially localized evidence. A promising solution is to integrate high-level RGB semantics with reconstruction-based compatibility cues, enabling the model to reason jointly about where suspicious artifacts occur, what they represent, and why they support the final forensic judgment.

LaP-Forensics (Latent-Pixel Forensics) complements RGB semantics with an explicit reconstruction-based compatibility signal for explainable image forensics. Specifically, a frozen Stable Diffusion v1.5 model performs deterministic DDIM inversion and reconstruction, providing a fixed reference against which the input image is compared. The resulting residual map characterizes local reconstruction compatibility: regions with lower residuals are more readily reconstructed by the reference model, whereas higher residuals indicate greater mismatch~\cite{wang2023dire,sha2024zerofake,song2021ddim}. Rather than treating this residual as a direct manipulation probability or source-attribution cue, LaP-Forensics encodes it jointly with RGB content as auxiliary forensic evidence. Building upon the \textit{Where--What--Why} reasoning paradigm~\cite{zhou2025where}, the model jointly produces structured forensic analyses and pixel-level artifact masks. To further align its outputs with the supplied evidence, we introduce a GRPO-based objective that rewards localization quality, structural compliance, and explicit references to the residual map. Since the text-side reward evaluates observable output properties rather than semantic entailment, it is formulated as evidence-reference regularization rather than a guarantee of explanation faithfulness.
Our main contributions are summarized as follows:
\begin{itemize}
    \item We propose LaP-Forensics, a dual-stream framework that integrates RGB semantics with DDIM-derived reconstruction residuals for image detection, reasoning, and artifact localization.

    \item We introduce a forensic \textit{Where--What--Why} protocol and a GRPO objective that jointly optimize localization quality, output structure, and evidence referencing.

    \item Extensive experiments on cross-generator detection and artifact localization validate the effectiveness of the residual stream and its key design choices.
\end{itemize}

\section{Related Work}
\label{sec:related}

\noindent\textbf{General Deepfake Detection.}
Image forensics has evolved from manipulation localization in RGB and noise domains~\cite{zhou2018richfeatures} to the open-world detection of fully synthetic images. Early methods exploited compression artifacts~\cite{rossler2019faceforensicspp}, blending boundaries~\cite{li2020facexray}, CNN fingerprints~\cite{wang2020cnnspot}, and frequency irregularities~\cite{qian2020f3net,luo2021highfrequency}. Recent methods instead emphasize cross-generator generalization through pretrained representations, inversion, and contrastive learning, as exemplified by UnivFD~\cite{ojha2023univfd}, FakeInversion~\cite{cazenavette2024fakeinversion}, FatFormer~\cite{liu2024fatformer}, and DRCT~\cite{chen2024drct}. Other approaches exploit patch-level discrepancies~\cite{yang2025all}, reconstruction residuals~\cite{shi2025face}, architectural priors~\cite{tan2024upsampling}, discrepancy learning~\cite{yang2025d3}, few-shot attribution~\cite{wu2025omnidfa}, region-grounded re-examination~\cite{ji2025zoom}, and high-resolution details~\cite{mu2025nopixel}. Despite strong detection performance, most provide only image-level scores or coarse localization, with limited semantic explanation~\cite{yan2025sanity,ji2025interpretable}.

\noindent\textbf{Vision--Language Reasoning and Explainable Forensics.}
Outside image forensics, language-conditioned visual systems use textual semantics to guide representation learning, as exemplified by motion-aware gait recognition with natural-language priors~\cite{wu2026language}. Multimodal agents also increasingly pair supervised fine-tuning with sequence-level reinforcement learning to optimize multi-step image and text search~\cite{wu2026promsa}. In image forensics, vision-language assistants such as MiniGPT-4~\cite{zhu2023minigpt4} and LLaVA~\cite{liu2023llava} have motivated models that jointly perform prediction, localization, and explanation. Representative systems include LEGION~\cite{kang2025legion}, SIDA~\cite{huang2025sida}, ManipShield~\cite{xu2025manipshield}, and related methods for face forensics and editing attribution~\cite{sun2025visuallinguistic,jiang2025edittrack,lin2025seeing}. However, these models often rely on visible semantic anomalies, such as distorted hands or implausible geometry. For visually polished forgeries, they may therefore generate persuasive explanations that are weakly grounded in explicit forensic evidence~\cite{ji2025interpretable,sun2025visuallinguistic}.

\noindent\textbf{Consistency-based Detection.}
Reconstruction-based forensics assesses whether an image is compatible with a learned reconstruction reference rather than relying solely on visible artifacts. DIRE~\cite{wang2023dire} introduced reconstruction errors for diffusion-image detection, while ZeroFake~\cite{sha2024zerofake}, JRC~\cite{yan2024jrc}, and DRCT~\cite{chen2024drct} further explored reconstruction discrepancies for generalized detection. More recently, studies have moved beyond surface artifacts to investigate shared latent forgery knowledge across generators and modalities~\cite{dou2026dna,dou2026beyond}. Different from these works, LaP-Forensics uses reconstruction residuals as an explicit multimodal cue for joint artifact localization and evidence-referenced language generation, and further evaluates the model's dependence on this cue through controlled interventions.

\section{Method}
\label{sec:method}
\subsection{Architecture Overview}
\begin{figure*}[t]
  \centering
\includegraphics[width=0.98\linewidth]{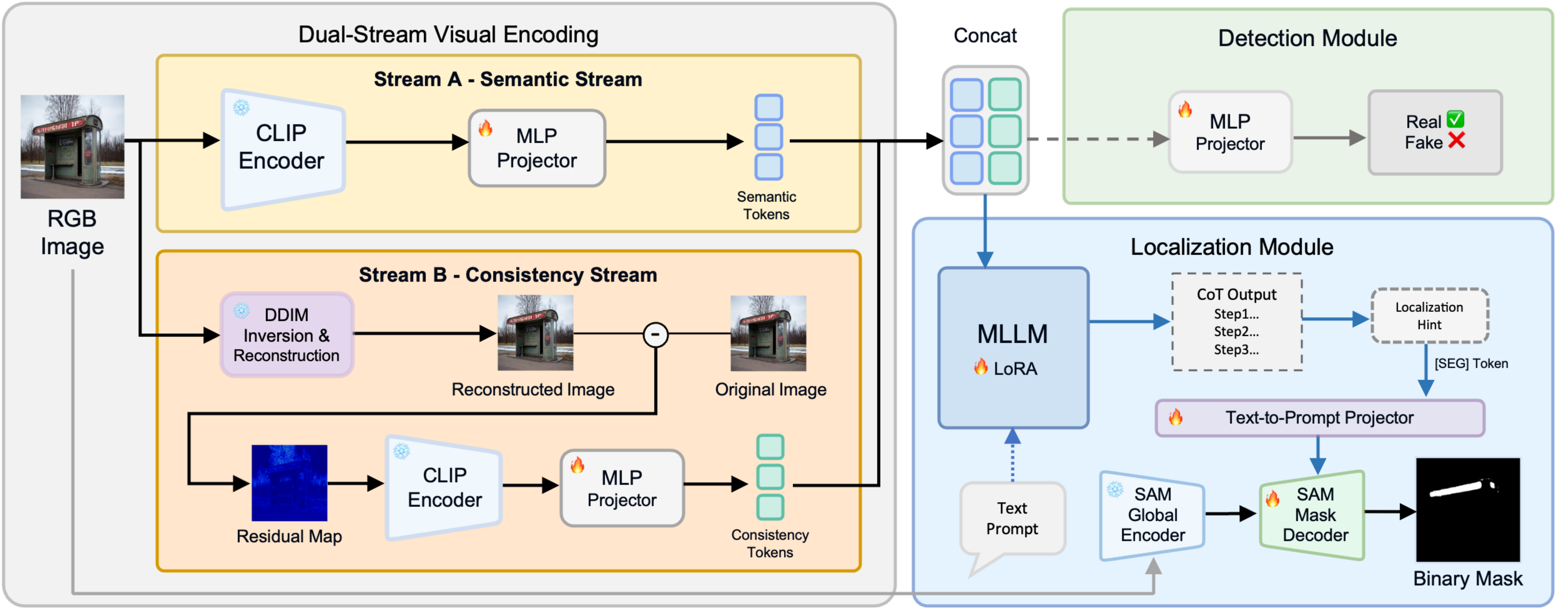}
  \caption{Overview of the dual-stream interface. RGB and DDIM-residual inputs share a frozen CLIP encoder but use independent projectors; their projected tokens are concatenated for the MLLM, and generated \texttt{[SEG]} states are decoded by SAM. The separately trained detector retains the same inputs but concatenates only the two \texttt{[CLS]} features before its MLP head.}
  \Description{Architecture diagram with an RGB semantic stream and a DDIM reconstruction-residual stream. A shared-weight frozen CLIP encoder and two independent MLP projectors produce the two token sequences. The localization branch concatenates the sequences for an MLLM and uses a SAM decoder to produce a mask. A separate detector concatenates stream-level class features and applies an MLP head.}
  \label{fig:overview}
\end{figure*}

\begin{figure}[t]
  \centering
\includegraphics[width=0.99\linewidth]{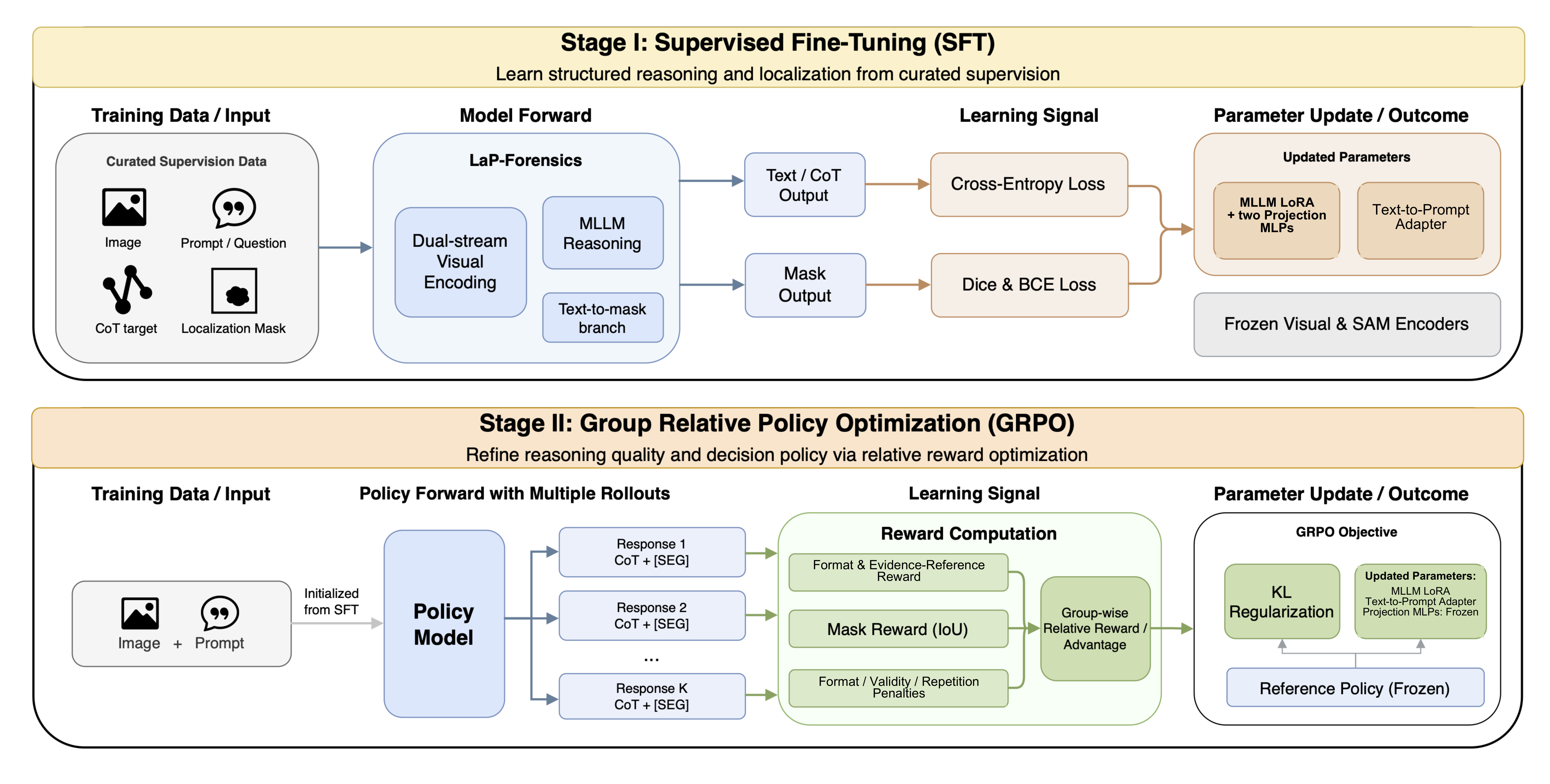}
  \vspace{-0.3cm}
  \caption{Two-stage optimization of the localization/reasoning model. Stage I jointly learns structured text generation and artifact segmentation. Stage II uses mask quality, output-format checks, residual-map references, and repetition control, with a clipped policy update and KL regularization to the frozen SFT reference policy. These observable reward terms encourage evidence-referencing outputs but do not directly verify the semantic truth of free-form text.}
  \label{fig:training_pipeline}
\end{figure}

LaP-Forensics is a two-stage training and alignment framework for explainable deepfake analysis. It combines a Multimodal Large Language Model (MLLM) with dual visual streams: an RGB semantic stream and a DDIM reconstruction-residual stream. The localization/reasoning model and the standalone detector evaluated in Section~\ref{sec:detection} use the same fixed DDIM residual construction but separate task heads and training objectives.
As illustrated in Figure~\ref{fig:overview}, an input image $\mathbf{x} \in \mathbb{R}^{H \times W \times 3}$ and its task-specific residual map are independently encoded and projected into a common token dimension. The full localization model feeds both token sequences to a LLaMA-2-7B language backbone \cite{touvron2023llama2}; a Segment Anything Model (SAM) \cite{kirillov2023sam} decoder converts generated \texttt{[SEG]} states into pixel masks. The standalone image detector instead concatenates the two stream-level class tokens and applies a lightweight MLP head. The reasoning model is optimized in two phases (Figure~\ref{fig:training_pipeline}): supervised fine-tuning (SFT) on curated forensic reasoning data followed by Group Relative Policy Optimization (GRPO) \cite{shao2024deepseekmath} for mask quality and structured evidence references.

\subsection{Dual-Stream Visual Representation}

\noindent\textbf{Semantic Feature Extraction.} To capture scene understanding and object-level semantics, we process the input image $\mathbf{x}$ using a shared-weight frozen CLIP-ViT-L/14 encoder $f_{\text{CLIP}}(\cdot)$ \cite{radford2021clip}. We extract patch-level features from the penultimate transformer layer:
\begin{equation}
\mathbf{Z}^{(s)} = \phi_{\text{proj}}(f_{\text{CLIP}}(\mathbf{x})) \in \mathbb{R}^{N_p \times d},
\end{equation}
where $N_p$ denotes the number of spatial patches and $\phi_{\text{proj}}(\cdot)$ is a two-layer MLP projector that aligns the CLIP feature dimension to the language model's hidden dimension $d$.

\noindent\textbf{Latent-Pixel Consistency Stream.} For localization and reasoning, we perform DDIM inversion \cite{song2021ddim} with a fixed $T=50$ schedule on a frozen Stable Diffusion v1.5 backbone. This frozen model provides a fixed reconstruction reference. The inversion maps the image latent $\mathbf{x}_0$ to terminal noise $\mathbf{z}_T$, and the deterministic reverse process yields $\hat{\mathbf{x}}$. We compute the element-wise compatibility residual
\begin{equation}
\mathbf{R} = |\mathbf{x} - \hat{\mathbf{x}}|.
\end{equation}
The residual is defined relative to the frozen reconstruction reference rather than as a direct manipulation probability. Lower values in $\mathbf{R}$ indicate regions reconstructed more readily by that reference, whereas higher values indicate larger mismatch. Such mismatch may expose synthetic or edited structures, but it can also respond to benign texture, high-frequency detail, compression, resizing, or domain shift. We therefore neither threshold $\mathbf{R}$ into a standalone manipulation mask nor use it to identify the source generator.

Instead, the spatial residual pattern is encoded and interpreted jointly with the RGB stream. This lets the downstream model combine semantic context with reconstruction compatibility rather than treating residual magnitude alone as an authenticity decision. For display, the Latent-Pixel Consistency Map uses a fixed grayscale rendering of $\mathbf{R}$: darker regions denote lower residual values and brighter regions denote larger mismatch. The display transformation is used only for interpretation and is not a calibrated probability map. We resize the input to $512 \times 512$, compute $\mathbf{R}$, and encode it with the same shared-weight frozen CLIP vision encoder:
\begin{equation}
\mathbf{Z}^{(c)} = \psi_{\text{proj}}(f_{\text{CLIP}}(\mathbf{R})) \in \mathbb{R}^{N_p \times d},
\end{equation}
where $\psi_{\text{proj}}(\cdot)$ is a dedicated two-layer MLP, independent of the semantic projector $\phi_{\text{proj}}$, and maps residual features into the language-model dimension while preserving a separate representation for reconstruction compatibility.

\noindent\textbf{Feature Fusion and Image-Level Classification.} The two streams share the frozen CLIP weights but do not share a projector: RGB and residual features pass through $\phi_{\text{proj}}$ and $\psi_{\text{proj}}$, respectively. For region-level reasoning, their projected patch tokens are concatenated in the sequence dimension with the text prompt,
$\mathbf{X} = [\mathbf{Z}^{(s)}, \mathbf{Z}^{(c)}, \mathbf{t}_{\text{prompt}}]$, so the MLLM can attend to both semantic context and the reconstruction signal.

The standalone detector follows a separate training and inference path. It applies the shared-weight frozen CLIP encoder separately to the RGB image and DDIM residual $\mathbf{R}$, concatenates the two resulting $\texttt{[CLS]}$ vectors, $\mathbf{c}^{(s)}$ and $\mathbf{c}^{(c)}$, and applies a two-layer MLP: $\hat{\mathbf{y}}_{\text{det}} = \mathrm{MLP}([\mathbf{c}^{(s)}; \mathbf{c}^{(c)}])$. Only this detection head is optimized. Thus Table~\ref{tab:detection_results} evaluates the final RGB+DDIM-residual detector, whereas the localization/reasoning branch additionally uses token-level fusion, the language backbone, and SAM.

\subsection{Supervised Fine-Tuning (Stage I)}

\noindent\textbf{Chain-of-Thought Data Construction.}
To connect visual observations with explicit forensic evidence, we organize each training sample into a three-step Chain-of-Thought (CoT). The first step identifies observable semantic or visual anomalies, such as distorted textures or inconsistent illumination. The second step relates these observations to high- or low-response regions in the Latent-Pixel Consistency Map. The final step integrates the preceding evidence to produce an authenticity judgment and appends a \texttt{[SEG]} token to trigger mask generation for the corresponding artifact region.

\noindent\textbf{Joint Training Objective.} During SFT, we train LoRA adapters \cite{hu2021lora} on the LLaMA-2-7B backbone together with the semantic projector $\phi_{\text{proj}}$, consistency projector $\psi_{\text{proj}}$, and text-to-prompt adapter $\phi_{\text{tp}}$, while keeping CLIP, DDIM, and SAM frozen. We jointly optimize language modeling and artifact segmentation. For segmentation, the hidden state corresponding to the $\texttt{[SEG]}$ token, $\mathbf{h}_{\texttt{[SEG]}}$, is processed by $\phi_{\text{tp}}$ to produce a prompt embedding $\mathbf{p} \in \mathbb{R}^{256}$. This embedding guides the frozen SAM decoder \cite{kirillov2023sam} to output a predicted mask $\hat{\mathbf{M}}$. 
The comprehensive SFT loss $\mathcal{L}_{\text{SFT}}$ integrates cross-entropy for text generation ($\mathcal{L}_{\text{CE}}$) with Dice ($\mathcal{L}_{\text{Dice}}$) and Binary Cross-Entropy ($\mathcal{L}_{\text{BCE}}$) losses for precise mask supervision:
\begin{equation}
\mathcal{L}_{\text{SFT}} = \lambda_{\text{ce}} \mathcal{L}_{\text{CE}} + \lambda_{\text{dice}} \mathcal{L}_{\text{Dice}} + \lambda_{\text{bce}} \mathcal{L}_{\text{BCE}}.
\end{equation}

\subsection{Outcome-Driven Alignment (Stage II)}
While SFT optimizes token-level likelihood, it does not directly optimize the mask produced from a generated \texttt{[SEG]} state. We therefore introduce a Group Relative Policy Optimization (GRPO) stage \cite{shao2024deepseekmath} that combines mask accuracy with observable text-structure constraints.

\noindent\textbf{Optimization Strategy.} GRPO is a memory-efficient algorithm that computes advantages relative to a sampled group, bypassing the need for a separate critic network. For a given prompt $\mathbf{x}$, we sample $G=4$ completions $\{\mathbf{y}^{(1)}, \ldots, \mathbf{y}^{(G)}\}$. The relative advantage $\hat{A}^{(i)}$ for completion $i$ is calculated as:
\begin{equation}
\hat{A}^{(i)} = \frac{R(\mathbf{x}, \mathbf{y}^{(i)}) - \text{mean}(\{R(\mathbf{x}, \mathbf{y}^{(j)})\}_{j=1}^G)}{\text{std}(\{R(\mathbf{x}, \mathbf{y}^{(j)})\}_{j=1}^G) + \epsilon}.
\end{equation}
For clarity, we write the group-relative policy term as
\begin{equation}
\mathcal{L}_{\text{GRPO}}
=-\mathbb{E}_{\mathbf{x},\{\mathbf{y}^{(i)}\}}
\left[
\frac{1}{G}\sum_{i=1}^{G}
\hat{A}^{(i)}
\log \pi_\theta(\mathbf{y}^{(i)}\mid\mathbf{x})
\right].
\end{equation}
The implementation additionally applies PPO-style policy-ratio clipping and KL regularization relative to a frozen reference policy initialized from the SFT checkpoint; the displayed equation isolates the group-relative term. During GRPO, the shared CLIP encoder, DDIM module, SAM, and both SFT-trained visual projectors remain frozen. Only the LoRA adapters and text-to-prompt adapter receive gradient updates.

\noindent\textbf{Reward Function Design.}\label{sec:reward-design} The reward is $R_{\text{total}} = \alpha_{\text{mask}} R_{\text{mask}} + \alpha_{\text{format}} R_{\text{format}}$, with $\alpha_{\text{mask}} = 0.7$ and $\alpha_{\text{format}} = 0.6$. The two terms measure mask overlap and observable output compliance, respectively.

\textit{Mask Reward ($R_{\text{mask}}$):} Evaluates segmentation accuracy against the ground-truth mask $\mathbf{M}^{*}$ using Intersection over Union (IoU), adding a bonus for high-quality predictions:
\begin{equation}
R_{\text{mask}} = \text{IoU}(\hat{\mathbf{M}}, \mathbf{M}^{*}) + \mathbb{1}[\text{IoU} > 0.7] \cdot 0.5.
\end{equation}

\textit{Format and Evidence-Reference Reward ($R_{\text{format}}$):} This term rewards the required step structure, the presence of \texttt{[SEG]}, explicit references to the consistency map, and a valid conclusion, while penalizing invalid character sequences and repetition:
\begin{equation}
R_{\text{format}} = \max(-3.0, \min(1.0, R_{\text{pos}} - P_{\text{invalid}} - P_{\text{rep}})).
\end{equation}
Here, $R_{\text{pos}}$ is computed from structural markers, evidence keywords, and conclusion markers. Missing \texttt{[SEG]} receives a separate $-2.0$ penalty; $P_{\text{invalid}}$ is triggered by a low ASCII-validity ratio and $P_{\text{rep}}$ by repetitive sentences. None of these rules evaluates whether a textual claim is semantically entailed by the consistency map. We therefore interpret this term as structural and evidence-reference regularization rather than a faithfulness reward.

\section{Experiments}\label{sec:experiments}

\subsection{Datasets}

\noindent\textbf{Primary Benchmark (SynthScars).} We conduct our experiments primarily on SynthScars \cite{kang2025legion}, a recently released expert-annotated dataset designed for explainable AI-generated image forensics. Comprising 12,236 fully synthetic images generated by modern diffusion models, SynthScars provides pixel-level segmentation masks, fine-grained textual explanations, and artifact-type labels. The images are distributed across four content categories (Human, Object, Animal, and Scene) and encompass diverse forgery patterns, including structural anomalies, texture inconsistencies, and semantic implausibilities. This rich annotation makes it an ideal foundation for multimodal reasoning and localization tasks.

\noindent\textbf{Chain-of-Thought Data Construction.} To support the structured \textit{Where, What, Why} protocol, we curate a subset of SynthScars through a semi-automatic pipeline. We first compute Stable Diffusion v1.5 residual maps with a fixed $T=50$ DDIM schedule. A vision-language model (Qwen3-VL-Plus\footnote{Official Qwen3-VL source: \url{https://github.com/QwenLM/Qwen3-VL}.}) then generates structured fields conditioned on the RGB image, consistency map, and original expert annotations. Human reviewers screen and edit the generated annotations for evident factual and formatting errors before the fields are converted into three-step training targets with the mandatory \texttt{[SEG]} marker appended to the inference span. 

\noindent\textbf{Curated Split Statistics.} The resulting curated data contain 3,673 unique images and 4,857 CoT annotations, as a single image may contain multiple independently annotated regions. We partition the annotations into 4,611 training entries and 246 test entries. All annotations associated with the same source image are assigned to a single split, so the training and test partitions are image-disjoint. Controlled CoT ablations use this curated test split; they are separate from the full official SynthScars evaluation in Table~\ref{tab:main_results}.

\noindent\textbf{Evaluation Protocol.} SFT and GRPO use the same 4,611-entry CoT training split. The baseline-ranking results in Table~\ref{tab:main_results} use the full official SynthScars evaluation protocol; the curated test entries are used only for controlled CoT and component ablations. Scores should therefore be compared within a table, not across the official and curated protocols. LOKI and RichHF are not used in CoT training and provide additional out-of-domain evaluations.

\begin{table*}[t]
\centering
\resizebox{\textwidth}{!}{%
\setlength{\tabcolsep}{2pt}
\begin{tabular}{l|cc|cc|cc|cc|cc|cc}
\toprule
\multirow{3}{*}{\textbf{Method}}
& \multicolumn{8}{c|}{\textbf{SynthScars}}
& \multicolumn{2}{c|}{\textbf{LOKI}}
& \multicolumn{2}{c}{\textbf{RichHF}} \\
\cline{2-13}
& \multicolumn{2}{c|}{Object}
& \multicolumn{2}{c|}{Animal}
& \multicolumn{2}{c|}{Human}
& \multicolumn{2}{c|}{Scene}
& \multirow{2}{*}{\textbf{mIoU}}
& \multirow{2}{*}{\textbf{F1}}
& \multirow{2}{*}{\textbf{mIoU}}
& \multirow{2}{*}{\textbf{F1}} \\
\cline{2-9}
& \textbf{mIoU} & \textbf{F1}
& \textbf{mIoU} & \textbf{F1}
& \textbf{mIoU} & \textbf{F1}
& \textbf{mIoU} & \textbf{F1}
& & & & \\
\midrule
HiFi-Net \cite{guo2023hifinet}
& 43.74 & 0.45 & 45.28 & 0.03 & 46.21 & 0.84
& 45.90 & 0.04 & 39.60 & 2.41 & 44.96 & 0.39 \\

TruFor \cite{guillaro2023trufor}
& 46.99 & 14.82 & 48.45 & 17.57 & 49.02 & 15.43
& 48.93 & 12.64 & 46.55 & 16.70 & 48.41 & 18.03 \\

PAL4VST \cite{zhang2023pal4vst}
& 50.46 & 19.25 & 52.55 & 21.61 & 59.18 & 35.70
& 52.55 & 19.14 & 47.34 & 11.58 & 49.88 & 14.78 \\

LISA-v1-7B \cite{lai2024lisa}
& 35.49 & 23.70 & 32.44 & 18.77 & 34.11 & 17.50
& 37.56 & 18.31 & 31.10 & 9.29 & 35.90 & 21.94 \\

InternVL2-8B \cite{chen2024internvl}
& 41.08 & 13.36 & 41.22 & 7.83 & 41.21 & 3.91
& 41.68 & 7.55 & 42.03 & 10.06 & 39.90 & 9.58 \\

Qwen2-VL-72B \cite{wang2024qwen2vl}
& 33.89 & 23.25 & 32.46 & 21.98 & 26.92 & 14.75
& 39.00 & 18.17 & 26.62 & 20.99 & 27.58 & 19.02 \\

LEGION \cite{kang2025legion}
& \underline{54.62} & 29.90
& 54.52 & 27.43
& \underline{60.82} & \underline{39.44}
& \underline{53.67} & 24.51
& \underline{48.66} & 16.71
& \underline{50.07} & 17.41 \\

Gemini-3-Flash (ours eval.)
& 52.69 & \underline{48.14}
& \textbf{73.92} & \textbf{75.99}
& 48.46 & 29.67
& 47.36 & \underline{50.75}
& 40.22 & \underline{25.99}
& 40.25 & \underline{26.89} \\
\midrule
\textbf{Ours}
& \textbf{55.40} & \textbf{60.20}
& \underline{55.30} & \underline{62.10}
& \textbf{61.40} & \textbf{66.00}
& \textbf{54.20} & \textbf{59.11}
& \textbf{51.40} & \textbf{37.91}
& \textbf{52.27} & \textbf{36.61} \\
\bottomrule
\end{tabular}%
}
\vspace{-0.1cm}
\caption{Artifact localization performance, measured by mIoU and F1, under the official SynthScars protocol and on the LOKI and RichHF benchmarks. This comparison is conducted independently of the 246-entry curated CoT test split used for controlled ablations. Published baseline results are taken from the SynthScars benchmark, whereas Gemini-3-Flash is evaluated by us under the same protocol.}
\label{tab:main_results}
\end{table*}

\subsection{Implementation Details}

We implement LaP-Forensics with PyTorch and TRL. In the localization/reasoning model, the shared-weight CLIP vision encoder, DDIM inversion--reconstruction module, and SAM image encoder/decoder remain frozen. The standalone detector is a distinct task head over the same frozen DDIM reconstruction module and shared-weight frozen CLIP encoder.

\noindent\textbf{Supervised Fine-Tuning (SFT).} In the supervised fine-tuning stage, we train the model on the SynthScars-CoT dataset for 5 epochs using the AdamW optimizer with a learning rate of $\mathbf{2 \times 10^{-5}}$, $\beta = (0.9, 0.95)$, and no weight decay. We apply a linear warmup schedule with 100 warmup steps, followed by linear decay. Training is performed in bf16 precision using DeepSpeed ZeRO-2 on four RTX 4090 GPUs. The per-GPU batch size is set to 7, resulting in an effective batch size of 28 without gradient accumulation. In total, the SFT stage runs for approximately 820 optimization steps. Low-Rank Adaptation (LoRA) \cite{hu2021lora} is applied to the language model backbone with rank $r = 8$, scaling factor $\alpha = 16$, and dropout rate $0.05$. Following prior practice, LoRA adapters are injected only into the query and value projection matrices (\texttt{q\_proj},\allowbreak\ \texttt{v\_proj}) of the transformer layers. SFT updates these LoRA adapters, both visual projectors $\phi_{\text{proj}}$ and $\psi_{\text{proj}}$, and the text-to-prompt adapter $\phi_{\text{tp}}$; the shared CLIP encoder, DDIM inversion--reconstruction module, and SAM encoder-decoder remain frozen.

\noindent\textbf{Reinforcement Learning with GRPO.}
After SFT converges at approximately 820 optimization steps, we further train the model for three epochs on the same SynthScars-CoT training split using AdamW with a learning rate of $1.0 \times 10^{-5}$. The per-device batch size is 4 with four gradient-accumulation steps. For each prompt, we sample $G=4$ completions with a temperature of 0.8 and a maximum generation length of 512 tokens. During this stage, only the LoRA and text-to-prompt adapters are updated, while the shared CLIP encoder, DDIM module, visual projectors, and SAM components remain frozen. Optimization employs the clipped GRPO objective with KL regularization against the frozen SFT reference policy. As detailed in Section~\ref{sec:reward-design}, the mask and format/evidence-reference rewards are weighted by 0.7 and 0.6, respectively, with an additional 0.5 bonus for predictions whose mask IoU exceeds 0.7. This stage jointly improves artifact localization and structured evidence referencing, but does not guarantee the semantic faithfulness of the generated explanations.

\noindent\textbf{Standalone Detection Training.} For image-level detection, we train only a binary head over RGB and residual features extracted by the shared-weight frozen CLIP backbone using cross-entropy. The forensic stream uses the same frozen Stable Diffusion v1.5 DDIM inversion--reconstruction pipeline with $T=50$ and computes $\mathbf{R}=|\mathbf{x}-\hat{\mathbf{x}}|$. Following the LEGION protocol \cite{kang2025legion}, the detector is trained on ProGAN and evaluated on UniversalFakeDetect. We use AdamW at $\mathbf{1 \times 10^{-4}}$, cosine learning-rate decay, bf16 precision, and concatenation of semantic and residual \texttt{[CLS]} features.

\subsection{Baselines and Evaluation Protocol}

We evaluate artifact localization under two deliberately separated protocols. Table~\ref{tab:main_results} follows the full official SynthScars evaluation and additionally reports LOKI and RichHF. The comparison includes forensic localization networks such as HiFi-Net, TruFor, and PAL4VST; general or multimodal segmentation models such as LISA, InternVL2, and Qwen2-VL; and the forensic reasoning model LEGION. Published baseline values are taken from the SynthScars benchmark rather than reimplemented, while Gemini-3-Flash is evaluated by us under the same official protocol.\footnote{Official Gemini 3 Flash model documentation: \url{https://ai.google.dev/gemini-api/docs/models/gemini-3-flash-preview}.} These rows are used for benchmark-level comparison.
The controlled component, cue-construction, reward, and counterfactual analyses in Section~\ref{sec:ablation} instead use the image-disjoint SynthScars-CoT split with 246 test entries. They compare variants within a fixed internal setting and should not be compared numerically with the official-split results in Table~\ref{tab:main_results}.
For image-level detection, we follow the UniversalFakeDetect protocol used by LEGION \cite{kang2025legion}: the LaP-Forensics detector is trained on ProGAN and evaluated separately on seven generator families. Published detector rows are taken from the LEGION benchmark report. DRCT is our evaluation of the official DRCT-2M checkpoint under the same folder-level test protocol, and the LaP-Forensics row corresponds to the separately trained RGB+DDIM-residual detector described in Section~\ref{sec:method}. Because the compared checkpoints use method-specific objectives and training sources, Table~\ref{tab:detection_results} is a benchmark comparison rather than an isolated architectural ablation.

\subsection{Detection Performance}\label{sec:detection}

\begin{figure}[t]
  \centering
  \includegraphics[width=\columnwidth]{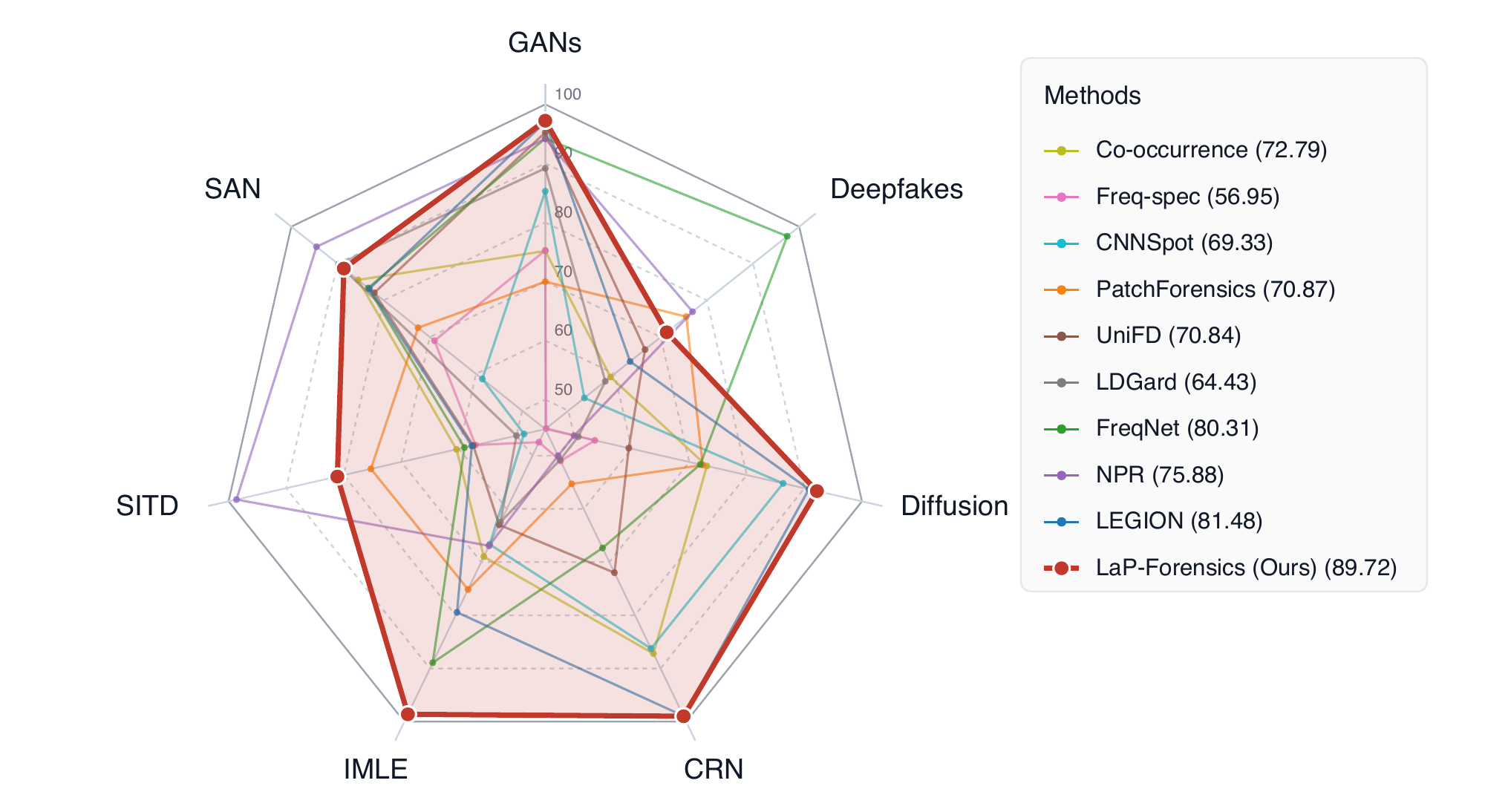}
  \caption{Family-level detection accuracy on UniversalFakeDetect, complementing Table~\ref{tab:detection_results}. The LaP-Forensics curve corresponds to the RGB+DDIM-residual detector.}
  \Description{A radar plot comparing detection accuracy across seven UniversalFakeDetect benchmark families for multiple methods, including the RGB+DDIM-residual LaP-Forensics detector.}
  \label{fig:detection_radar}
\end{figure}

Table~\ref{tab:detection_results} reports accuracy across the seven UniversalFakeDetect families, and Figure~\ref{fig:detection_radar} visualizes the same profile. LaP-Forensics obtains 97.23\% on GANs, 71.30\% on Deepfakes, 92.18\% on Diffusion, 98.98\% on CRN, 98.62\% on IMLE, 81.11\% on SITD, and 88.62\% on SAN. It gives the highest reported accuracy in the GANs, Diffusion, CRN, and IMLE columns. On Deepfakes, SITD, and SAN, specialized baselines remain stronger, illustrating that no evaluated detector dominates every generator family.
Relative to LEGION, LaP-Forensics is nearly tied on GANs (+0.22 points) and CRN (+0.05), while the largest differences occur on IMLE (+19.18) and SITD (+23.35). These family-level differences are descriptive and are not statistical significance estimates. The official DRCT-2M checkpoint is also lower in this evaluation, although the difference narrows to 0.72 points on SAN and 4.17 on SITD and is larger on the remaining families. This official-checkpoint comparison does not isolate the residual representation because the methods use different training objectives and checkpoints.
The resulting profile is not confined to diffusion-generated images: the RGB-plus-reconstruction representation also transfers numerically to GAN, CRN, and IMLE families under the evaluated protocol. We interpret the DDIM residual only as a fixed-reference compatibility cue. Performance on GAN or CRN images neither identifies their source mechanism nor implies that those generators follow a diffusion trajectory.
Table~\ref{tab:detection_subset_results} additionally reports LaP-Forensics on selected generator subsets from MSCOCOAI~\cite{roy2026mscocoai} and OpenSDI~\cite{wang2025opensdi}. These results complement the UniversalFakeDetect evaluation but do not constitute a comprehensive assessment of open-world distribution shift or post-processing reliability.

\begin{table}[t]
\centering
\scriptsize
\setlength{\tabcolsep}{3.5pt}
\resizebox{\columnwidth}{!}{%
\begin{tabular}{lccccccc}
\toprule
\textbf{Method} & \textbf{GANs} & \textbf{Deepfakes} & \textbf{Diffusion} & \textbf{CRN} & \textbf{IMLE} & \textbf{SITD} & \textbf{SAN} \\
\midrule
Co-occurrence \cite{nataraj2019cooccurrence} & 75.17 & 59.14 & 73.06 & \underline{87.21} & 68.98 & 60.42 & 85.53 \\
Freq-spec \cite{zhang2019freqspec} & 75.28 & 45.18 & 53.61 & 50.98 & 47.46 & 57.12 & 69.00 \\
CNNSpot \cite{wang2020cnnspot} & 85.29 & 53.47 & 86.31 & 86.26 & 66.67 & 48.69 & 58.63 \\
PatchForensics \cite{chai2020patchfor} & 69.97 & 75.54 & 72.33 & 55.30 & 75.14 & 75.28 & 72.54 \\
UniFD \cite{ojha2023univfd} & 95.25 & 66.60 & 59.50 & 72.00 & 63.00 & 57.50 & 82.02 \\
LDGard \cite{tan2023ldgard} & 89.17 & 58.00 & 50.74 & 50.78 & 62.50 & 50.00 & \underline{89.79} \\
FreqNet \cite{tan2024freqnet} & 94.23 & \textbf{97.40} & 71.92 & 67.35 & \underline{88.92} & 59.04 & 83.34 \\
NPR \cite{tan2024upsampling} & 94.16 & \underline{76.89} & 50.00 & 50.00 & 66.94 & \textbf{98.63} & \textbf{94.54} \\
LEGION \cite{kang2025legion} & \underline{97.01} & 63.37 & \underline{90.78} & \underline{98.93} & 79.44 & 57.76 & 83.10 \\
DRCT \cite{chen2024drct} & 56.66 & 57.10 & 78.33 & 50.06 & 51.18 & 76.94 & 87.90 \\
\midrule
\textbf{LaP-Forensics (Ours)} & \textbf{97.23} & 71.30 & \textbf{92.18} & \textbf{98.98} & \textbf{98.62} & \underline{81.11} & \underline{88.62} \\
\bottomrule
\end{tabular}
}
\caption{Synthetic-image detection accuracy (\%) on UniversalFakeDetect. Published baseline numbers are taken from \cite{kang2025legion}; DRCT is our evaluation of the official DRCT-2M checkpoint under the same folder protocol. The last row reports our separately trained RGB+DDIM-residual detector with \texttt{[CLS]} concatenation.}
\label{tab:detection_results}
\end{table}

\begin{table}[t]
\centering
\scriptsize
\setlength{\tabcolsep}{6pt}
\begin{tabular*}{\columnwidth}{
    @{\extracolsep{\fill}}lcc@{}}
\toprule
\textbf{Dataset} & \textbf{Generator} & \textbf{Accuracy (\%)} \\
\midrule
\multirow{2}{*}{MSCOCOAI}
    & DALL-E 3    & 80.89 \\
    & Midjourney  & 81.06 \\
\midrule
\multirow{5}{*}{OpenSDI}
    & SD1.5       & 83.02 \\
    & SD2.1       & 78.34 \\
    & SDXL        & 80.15 \\
    & SD3         & \textbf{88.56} \\
    & FLUX        & 81.71 \\
\bottomrule
\end{tabular*}
\caption{Generator-specific detection accuracy (\%) of LaP-Forensics on selected subsets of MSCOCOAI~\cite{roy2026mscocoai} and OpenSDI~\cite{wang2025opensdi}. This evaluation complements the UniversalFakeDetect results and is not intended as a comprehensive assessment under open-world or post-processing conditions.}
\label{tab:detection_subset_results}
\end{table}

\subsection{Qualitative Visualization}
\begin{figure}[t]
    \centering
\includegraphics[width=\linewidth]{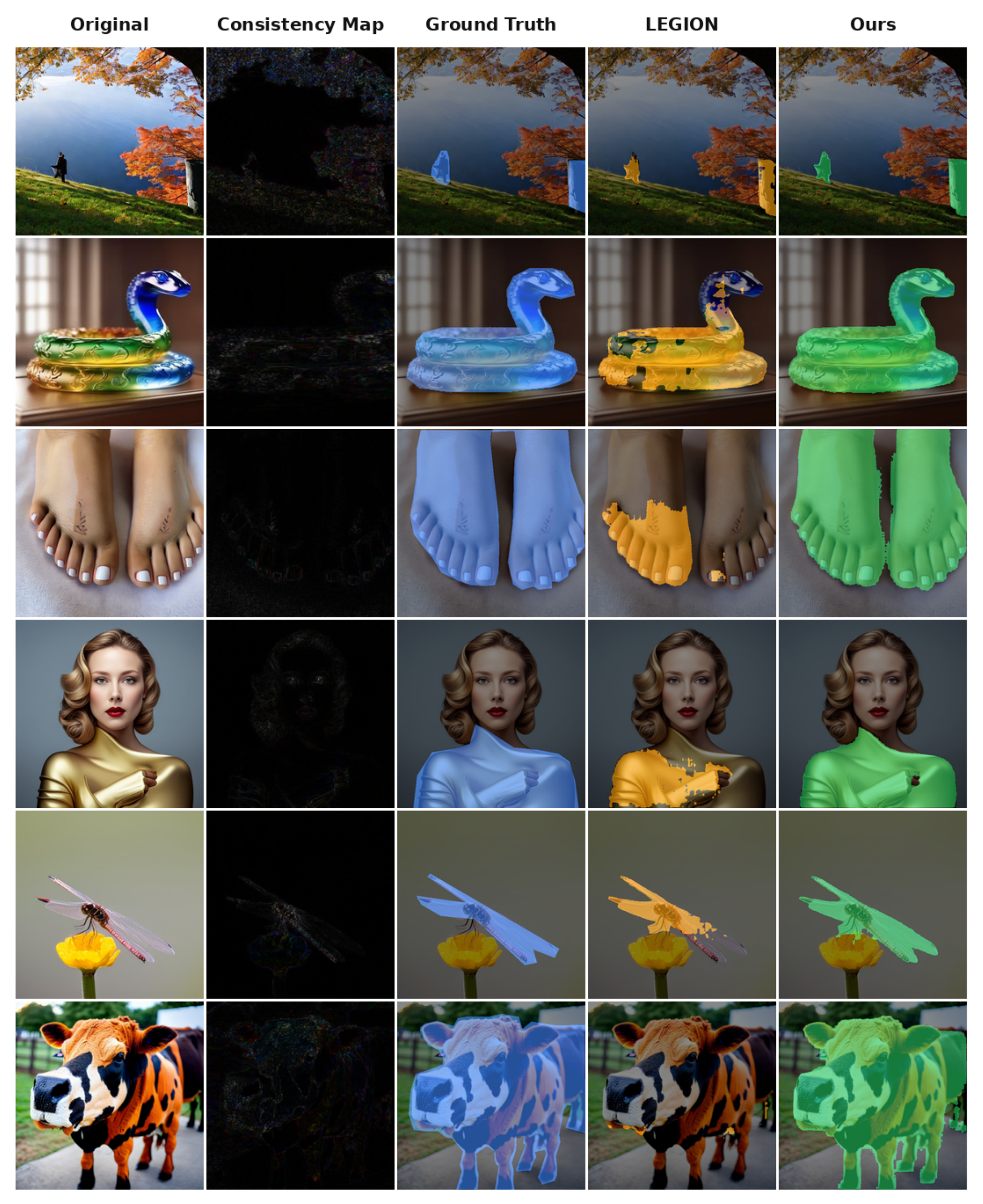}
\vspace{-0.2cm}
    \caption{Qualitative comparison against LEGION on six selected manipulated images. Each row shows the input image, consistency map, ground-truth edited region, LEGION mask, and LaP-Forensics mask. In these examples, our predictions more closely follow several fine boundaries and localized regions while reducing some scattered activations.}
    \label{fig:qualitative_vis}
\end{figure}

Figure~\ref{fig:qualitative_vis} compares six representative cases using the input image, consistency map, ground-truth mask, LEGION prediction, and LaP-Forensics prediction. In several examples, LaP-Forensics produces more spatially coherent masks that better cover the annotated regions, preserve thin or localized structures, and reduce disconnected activations. The comparison also highlights differences in boundary continuity and in the suppression of scattered responses. Notably, the consistency map provides complementary spatial cues when appearance-based anomalies are weak or distributed across multiple regions, helping the model associate reconstruction mismatch with semantic image content. However, some predictions remain conservative around subtle boundaries or small manipulated regions, indicating that weak residual responses can still lead to under-segmentation. These selected examples are intended to illustrate mask extent, boundary alignment, and fragmentation rather than establish uniform superiority; the quantitative results remain the primary basis for distribution-level comparison.

\subsection{Ablation Studies}\label{sec:ablation}
Table~\ref{tab:ablation} summarizes three controlled analyses under the same curated SynthScars-CoT protocol. Overall, the results show that the reconstruction residual is the most critical component, while structured CoT supervision and GRPO further improve localization and output alignment. Among the evaluated cue constructions, the multi-step DDIM residual clearly outperforms both RGB-only input and the single-step VAE residual, with $T=50$ providing the best balance among the tested inversion horizons. The reward ablation in Table~\ref{tab:reward_ablation} further demonstrates that mask, format, and evidence-reference objectives provide complementary supervision, while the counterfactual interventions in Table~\ref{tab:counterfactual_map} confirm that the predicted masks materially depend on the supplied consistency map. All results in this section are obtained using the curated internal split and should not be directly compared with the official benchmark results in Table~\ref{tab:main_results}.

\noindent\textbf{Model components.}
Removing the residual stream causes the largest performance drop, reducing mIoU from 72.19 to 61.83 and F1 from 63.62 to 41.71, which highlights the importance of reconstruction-based compatibility cues beyond RGB semantics alone. The substantially larger decrease in F1 further suggests that the residual stream is particularly important for preserving coherent artifact regions and reducing fragmented predictions. Removing the structured CoT supervision leads to a smaller but consistent decline to 69.72 mIoU and 59.18 F1, indicating that the \textit{Where--What--Why} reasoning structure also contributes to more accurate localization. This result implies that explicitly organizing observation, evidence analysis, and inference helps connect the generated \texttt{[SEG]} representation with the relevant forensic regions. Using SFT without the subsequent GRPO stage achieves 63.05/55.40, suggesting that supervised learning alone is insufficient to fully align the generated reasoning with the localization objective. In contrast, applying GRPO with only the mask reward further degrades performance to 55.46/37.91, showing that direct mask optimization without structural and evidence-reference constraints can produce unstable or poorly aligned outputs. The inferior mask-only result also indicates that optimizing a single outcome metric may disrupt the structured generation behavior learned during SFT. As further supported by Table~\ref{tab:reward_ablation}, adding format and evidence-reference rewards progressively improves both localization accuracy and CoT-format compliance. Overall, the best performance is achieved only when the residual stream, structured CoT supervision, and the complete GRPO reward are jointly employed, demonstrating that these components provide complementary rather than redundant benefits.

\noindent\textbf{Compatibility-cue construction.}
This ablation isolates the contribution of reconstruction-based information from the specific DDIM formulation. Compared with the RGB-only baseline, introducing a single-step VAE residual improves mIoU from 61.83 to 64.58 and F1 from 41.71 to 49.42, indicating that even a simple reconstruction cue provides complementary information beyond appearance semantics. Replacing the VAE residual with the $T=50$ DDIM residual further increases performance to 72.19 mIoU and 63.62 F1, corresponding to additional gains of 7.61 and 14.20 points, respectively. The larger improvement suggests that the multi-step inversion--reconstruction process yields a more informative spatial compatibility pattern than a single-step reconstruction. In particular, the DDIM residual appears to better expose localized mismatches that can be aligned with artifact boundaries during multimodal reasoning. These results support the use of the tested DDIM-based cue under the controlled setting, but do not imply that all source generators follow a diffusion trajectory or that the residual directly identifies the underlying generation mechanism.

\begin{table}[t]
\centering
\footnotesize
\setlength{\tabcolsep}{3.5pt}
\begin{tabular*}{\columnwidth}{
    @{\extracolsep{\fill}}lcc@{}}
\toprule
\textbf{Configuration} & \textbf{mIoU (\%)} & \textbf{F1 (\%)} \\
\midrule

\multicolumn{3}{@{}l}{\textit{Model components}} \\
Full model                  & \textbf{72.19} & \textbf{63.62} \\
w/o residual stream         & 61.83          & 41.71          \\
w/o CoT supervision         & 69.72          & 59.18          \\
SFT only                    & 63.05          & 55.40          \\
GRPO with mask reward only  & 55.46          & 37.91          \\

\midrule
\multicolumn{3}{@{}l}{\textit{Compatibility-cue construction}} \\
RGB input only              & 61.83          & 41.71          \\
Single-step VAE             & 64.58          & 49.42          \\
DDIM inversion ($T=50$)     & \textbf{72.19} & \textbf{63.62} \\

\midrule
\multicolumn{3}{@{}l}{\textit{DDIM inversion horizon}} \\
$T=20$                      & 69.89          & 56.72          \\
$T=50$                      & \textbf{72.19} & \textbf{63.62} \\
$T=100$                     & 60.50          & 48.77          \\
\bottomrule
\end{tabular*}
\caption{Ablation studies on the curated SynthScars-CoT test split, evaluating model components, compatibility-cue construction, and the DDIM inversion horizon. The best result in each group is highlighted in bold.}
\label{tab:ablation}
\end{table}
\noindent\textbf{DDIM inversion horizon.}
The inversion horizon exhibits a clear non-monotonic effect on localization performance. Increasing $T$ from 20 to 50 improves mIoU from 69.89 to 72.19 and F1 from 56.72 to 63.62, indicating that a moderate inversion depth produces a more informative compatibility signal. However, further increasing the horizon to $T=100$ substantially reduces performance to 60.50 mIoU and 48.77 F1. This degradation suggests that overly long inversion may introduce excessive reconstruction deviation or amplify benign image details, thereby weakening the spatial correspondence between the residual map and actual artifact regions. Among the evaluated settings, $T=50$ provides the best balance between residual sensitivity and localization reliability. Nevertheless, this result is specific to the current backbone and evaluation protocol and does not establish 50 steps as universally optimal across datasets, resolutions, or post-processing conditions.

\noindent\textbf{Per-term GRPO reward ablation.}
Table~\ref{tab:reward_ablation} examines the individual contributions of the mask, output-format, and evidence-reference rewards while keeping the SFT checkpoint and curated data fixed. Using the mask reward alone yields only 55.46 mIoU and 37.91 F1, indicating that direct localization supervision is insufficient to reliably align the generated \texttt{[SEG]} state with the desired output structure. Adding the format reward substantially improves performance to 70.84 mIoU and 61.15 F1, together with a CoT-Format-Rate of 94.6\%. The evidence-reference term also produces clear gains over the mask-only setting, reaching 68.32 mIoU, 58.47 F1, and 86.1\% format compliance. Combining all reward terms achieves the best overall results of 72.19 mIoU, 63.62 F1, and 97.5\% CoT-Format-Rate, showing that structural constraints and evidence referencing provide complementary supervision. CoT-Format-Rate measures compliance with the required three-step structure and the presence of the \texttt{[SEG]} marker, rather than the semantic truth or faithfulness of the generated explanation.

\begin{table}[t]
\centering
\scriptsize
\setlength{\tabcolsep}{4pt}
\begin{tabular*}{\columnwidth}{
    @{\extracolsep{\fill}}lccc@{}}
\toprule
\textbf{GRPO Reward Configuration}
& \textbf{mIoU (\%)}
& \textbf{F1 (\%)}
& \textbf{CoT-Fmt (\%)} \\
\midrule
Mask only
& 55.46 & 37.91 & 67.3 \\

Mask + format
& 70.84 & 61.15 & 94.6 \\

Mask + evidence reference
& 68.32 & 58.47 & 86.1 \\

Full reward
& \textbf{72.19}
& \textbf{63.62}
& \textbf{97.5} \\
\bottomrule
\end{tabular*}
\caption{Ablation of individual GRPO reward components on the curated SynthScars-CoT test split. CoT-Fmt denotes the rate of outputs satisfying the required CoT structure and formatting constraints. The best result in each column is highlighted in bold.}
\label{tab:reward_ablation}
\end{table}

\noindent\textbf{Counterfactual map intervention.}
To examine whether the model genuinely uses the consistency map, we perform two counterfactual interventions at inference: replacing the map with an all-zero input or with a donor map from another image. The evaluation covers all 246 examples in the curated test split using greedy decoding with seed 42, and predicted mask probabilities are binarized at 0.5 before computing mIoU@0.5. As reported in Table~\ref{tab:counterfactual_map}, removing the consistency signal reduces mIoU@0.5 from 0.721 to 0.604, while the masks predicted with the normal and zero maps achieve an average mutual IoU of only 0.46. Replacing the original map with a donor map yields a similar performance drop to 0.611 and causes 41.2\% of the predicted masks to shift toward high-response regions in the donor map. The zero-map intervention shows that performance cannot be recovered from RGB information alone, whereas the donor-map intervention further demonstrates sensitivity to the spatial content of the supplied consistency signal rather than merely the presence of a second input stream. Together, these results provide intervention-based evidence that the predicted localization masks materially depend on the consistency map. This analysis concerns spatial output dependence and does not establish the semantic faithfulness of the generated free-form explanations.

\noindent\textbf{Residual interpretation.}
The pixel-level DDIM residual can also respond to benign high-frequency structures, such as hair, woven fabrics, foliage, specular highlights, and resampling boundaries, even in the absence of manipulation. We therefore interpret the residual through region-level continuity, low-contrast spatial patterns, and boundary transitions in conjunction with the corresponding RGB content, rather than treating isolated high-response pixels as direct manipulation evidence. LaP-Forensics encodes the residual and RGB image using a shared-weight frozen vision backbone and fuses their representations, enabling the model to relate diffuse or fragmented residual patterns to semantic image structure. Nevertheless, the residual map is neither a calibrated manipulation-probability map nor an indicator of the source generator, and multimodal fusion cannot fully eliminate responses caused by nuisance textures or post-processing artifacts. The counterfactual experiments confirm that the predicted masks depend on the residual input, but they do not establish the validity of every residual activation or the semantic faithfulness of the generated explanations.

\begin{table}[t]
\centering
\scriptsize
\setlength{\tabcolsep}{4pt}
\begin{tabular*}{\columnwidth}{
    @{\extracolsep{\fill}}lc@{}}
\toprule
\textbf{Map Condition} & \textbf{mIoU@0.5} \\
\midrule
Normal map & \textbf{0.721} \\
Zero map   & 0.604 \\
Donor map  & 0.611 \\
\bottomrule
\end{tabular*}
\caption{Counterfactual interventions on the consistency map over all 246 examples in the curated SynthScars-CoT test split. Greedy decoding is used with seed 42, and predicted mask probabilities are binarized at a threshold of 0.5. This experiment evaluates the dependence of spatial predictions on the consistency map rather than the semantic faithfulness of free-form explanations.}
\label{tab:counterfactual_map}
\end{table}

\noindent\textbf{Limitations.}
The DDIM residual is tied to a fixed Stable Diffusion reference and may be affected by post-processing, subtle edits, and natural high-frequency textures, leading to weakened or spurious localization responses. Robustness under such degradations requires further systematic evaluation. Moreover, the text-side reward encourages evidence referencing but does not verify semantic faithfulness, and the curated CoT split is smaller than the official SynthScars test set; results across the two protocols should therefore not be directly compared.

\section{Conclusion}
In this paper, we proposed LaP-Forensics, a dual-stream multimodal framework for synthetic-image detection, explainable forensic reasoning, and artifact localization. Specifically, LaP-Forensics integrates RGB semantics with DDIM-derived reconstruction residuals, employs a structured \textit{Where--What--Why} protocol to generate forensic analyses and pixel-level masks, and introduces a GRPO objective to jointly optimize localization quality, output structure, and evidence referencing. Extensive experiments on UniversalFakeDetect and SynthScars demonstrate strong cross-generator detection performance and competitive artifact localization under the evaluated protocols. Further ablation studies on cue construction, inversion horizon, model components, reward terms, and counterfactual map interventions verify the effectiveness of the residual stream and the proposed alignment strategy. Nevertheless, robustness to post-processing operations and the semantic faithfulness of free-form explanations remain open challenges, which will be investigated in future work through degradation-aware training and more direct evidence-faithfulness evaluation.

\providecommand{\showeprint}[2][]{}
\renewcommand{\showeprint}[2][]{}

\bibliographystyle{ACM-Reference-Format}
\bibliography{references}

@inproceedings{cazenavette2024fakeinversion,
  author    = {Cazenavette, George and Sud, Avneesh and Leung, Thomas and Usman, Ben},
  title     = {{FakeInversion}: Learning to Detect Images from Unseen Text-to-Image Models by Inverting Stable Diffusion},
  booktitle = {2024 IEEE/CVF Conference on Computer Vision and Pattern Recognition (CVPR)},
  pages     = {10759--10769},
  year      = {2024},
  publisher = {IEEE Computer Society},
  address   = {Los Alamitos, CA, USA},
  doi       = {10.1109/CVPR52733.2024.01023}
}

@inproceedings{chen2024drct,
  author    = {Chen, Baoying and Zeng, Jishen and Yang, Jianquan and Yang, Rui},
  title     = {{DRCT}: Diffusion Reconstruction Contrastive Training towards Universal Detection of Diffusion Generated Images},
  booktitle = {Proceedings of the 41st International Conference on Machine Learning},
  series    = {Proceedings of Machine Learning Research},
  volume    = {235},
  pages     = {7621--7639},
  year      = {2024},
  publisher = {PMLR},
  address   = {Vienna, Austria},
  url       = {https://proceedings.mlr.press/v235/chen24ay.html}
}

@misc{huang2025sida,
  author        = {Huang, Zhenglin and Hu, Jinwei and Li, Xiangtai and He, Yiwei and Zhao, Xingyu and Peng, Bei and Wu, Baoyuan and Huang, Xiaowei and Cheng, Guangliang},
  title         = {{SIDA}: Social Media Image Deepfake Detection, Localization and Explanation with Large Multimodal Model},
  year          = {2025},
  eprint        = {2412.04292},
  archiveprefix = {arXiv},
  doi           = {10.48550/arXiv.2412.04292}
}

@misc{ji2025interpretable,
  author        = {Ji, Yikun and Yan, Hong and Lan, Jun and Zhu, Huijia and Wang, Weiqiang and Fan, Qi and Zhang, Liqing and Zhang, Jianfu},
  title         = {Interpretable and Reliable Detection of {AI}-Generated Images via Grounded Reasoning in {MLLMs}},
  year          = {2025},
  eprint        = {2506.07045},
  archiveprefix = {arXiv},
  doi           = {10.48550/arXiv.2506.07045}
}

@article{shen2024imagpose,
  title={Imagpose: A unified conditional framework for pose-guided person generation},
  author={Shen, Fei and Tang, Jinhui},
  journal={Advances in neural information processing systems},
  volume={37},
  pages={6246--6266},
  year={2024}
}

@inproceedings{shen2025imagdressing,
  title={Imagdressing-v1: Customizable virtual dressing},
  author={Shen, Fei and Jiang, Xin and He, Xin and Ye, Hu and Wang, Cong and Du, Xiaoyu and Li, Zechao and Tang, Jinhui},
  booktitle={Proceedings of the AAAI Conference on Artificial Intelligence},
  volume={39},
  number={7},
  pages={6795--6804},
  year={2025}
}

@inproceedings{shen2024advancing,
title={Advancing Pose-Guided Image Synthesis with Progressive Conditional Diffusion Models},
author={Fei Shen and Hu Ye and Jun Zhang and Cong Wang and Xiao Han and Yang Wei},
booktitle={The Twelfth International Conference on Learning Representations},
year={2024},
url={https://openreview.net/forum?id=rHzapPnCgT}
}

@inproceedings{shen2025boosting,
  title={Boosting consistency in story visualization with rich-contextual conditional diffusion models},
  author={Shen, Fei and Ye, Hu and Liu, Sibo and Zhang, Jun and Wang, Cong and Han, Xiao and Wei, Yang},
  booktitle={Proceedings of the AAAI Conference on Artificial Intelligence},
  volume={39},
  number={7},
  pages={6785--6794},
  year={2025}
}

@misc{ji2025zoom,
  author        = {Ji, Yikun and Hong, Yan and Deng, Bowen and Lan, Jun and Zhu, Huijia and Wang, Weiqiang and Zhang, Liqing and Zhang, Jianfu},
  title         = {Locate-Then-Examine: Grounded Region Reasoning Improves Detection of {AI}-Generated Images},
  year          = {2025},
  eprint        = {2510.04225},
  archiveprefix = {arXiv},
  doi           = {10.48550/arXiv.2510.04225}
}

@misc{jiang2025edittrack,
  author        = {Jiang, Zhengyuan and Zhang, Yuyang and Guo, Moyang and Gong, Neil Zhenqiang},
  title         = {{EditTrack}: Detecting and Attributing {AI}-assisted Image Editing},
  year          = {2025},
  eprint        = {2510.01173},
  archiveprefix = {arXiv},
  doi           = {10.48550/arXiv.2510.01173}
}

@misc{kang2025legion,
  author        = {Kang, Hengrui and Wen, Siwei and Wen, Zichen and Ye, Junyan and Li, Weijia and Feng, Peilin and Zhou, Baichuan and Wang, Bin and Lin, Dahua and Zhang, Linfeng and He, Conghui},
  title         = {{LEGION}: Learning to Ground and Explain for Synthetic Image Detection},
  year          = {2025},
  eprint        = {2503.15264},
  archiveprefix = {arXiv},
  doi           = {10.48550/arXiv.2503.15264}
}

@misc{lin2025seeing,
  author        = {Lin, Kaiqing and Yan, Zhiyuan and Chen, Ruoxin and Ye, Junyan and Zhang, Ke-Yue and Zhou, Yue and Jin, Peng and Li, Bin and Yao, Taiping and Ding, Shouhong},
  title         = {Seeing Before Reasoning: A Unified Framework for Generalizable and Explainable Fake Image Detection},
  year          = {2025},
  eprint        = {2509.25502},
  archiveprefix = {arXiv},
  doi           = {10.48550/arXiv.2509.25502}
}

@article{dou2026dna,
  title={DNA: Uncovering Universal Latent Forgery Knowledge},
  author={Dou, Jingtong and Shi, Chuancheng and Wang, Yemin and Guo, Shiming and Yi, Anqi and Wu, Wenhua and Zhang, Li and Shen, Fei and Chua, Tat-Seng},
  journal={arXiv preprint arXiv:2601.22515},
  year={2026}
}

@article{dou2026beyond,
  title={Beyond surface artifacts: Capturing shared latent forgery knowledge across modalities},
  author={Dou, Jingtong and Shi, Chuancheng and Wang, Jian and Shen, Fei and Wang, Zhiyong and Chua, Tat-Seng},
  journal={arXiv preprint arXiv:2604.07763},
  year={2026}
}

@article{shen2025imaggarment,
  title={IMAGGarment-1: Fine-Grained Garment Generation for Controllable Fashion Design},
  author={Shen, Fei and Yu, Jian and Wang, Cong and Jiang, Xin and Du, Xiaoyu and Tang, Jinhui},
  journal={arXiv preprint arXiv:2504.13176},
  year={2025}
}

@misc{mu2025nopixel,
  author        = {Mu, Lianrui and Zou, Xingze and Bai, Jianhong and Hu, Jiaqi and Zheng, Wenjie and Ye, Jiangnan and Zhuang, Jiedong and Ali, Mudassar and Wang, Jing and Hu, Haoji},
  title         = {No Pixel Left Behind: A Detail-Preserving Architecture for Robust High-Resolution {AI}-Generated Image Detection},
  year          = {2025},
  eprint        = {2508.17346},
  archiveprefix = {arXiv},
  doi           = {10.48550/arXiv.2508.17346}
}

@misc{roy2026mscocoai,
  author        = {Roy, Rajarshi and Aziz, Ashhar and Bajpai, Shashwat and Imanpour, Nasrin and Singh, Gurpreet and Biswas, Shwetangshu and Wanaskar, Kapil and Patwa, Parth and Ghosh, Subhankar and Dixit, Shreyas and Pal, Nilesh Ranjan and Rawte, Vipula and Garimella, Ritvik and Das, Amitava and Sheth, Amit and Jena, Gaytri and Sharma, Vasu and Reganti, Aishwarya Naresh and Jain, Vinija and Chadha, Aman},
  title         = {A Comprehensive Dataset for Human vs. {AI} Generated Image Detection},
  year          = {2026},
  eprint        = {2601.00553},
  archiveprefix = {arXiv},
  doi           = {10.48550/arXiv.2601.00553}
}

@inproceedings{sha2024zerofake,
  author    = {Sha, Zeyang and Tan, Yicong and Li, Mingjie and Backes, Michael and Zhang, Yang},
  title     = {{ZeroFake}: Zero-Shot Detection of Fake Images Generated and Edited by Text-to-Image Generation Models},
  booktitle = {Proceedings of the 2024 ACM SIGSAC Conference on Computer and Communications Security},
  pages     = {4852--4866},
  year      = {2024},
  publisher = {Association for Computing Machinery},
  address   = {New York, NY, USA},
  doi       = {10.1145/3658644.3690297}
}

@article{shi2025face,
  author  = {Shi, Zenan and Liu, Wenyu and Chen, Haipeng},
  title   = {Face Reconstruction-Based Generalized Deepfake Detection Model with Residual Outlook Attention},
  journal = {ACM Transactions on Multimedia Computing, Communications, and Applications},
  volume  = {21},
  number  = {4},
  pages   = {1--19},
  year    = {2025},
  doi     = {10.1145/3686162}
}

@misc{wang2023dire,
  author        = {Wang, Zhendong and Bao, Jianmin and Zhou, Wengang and Wang, Weilun and Hu, Hezhen and Chen, Hong and Li, Houqiang},
  title         = {{DIRE} for Diffusion-Generated Image Detection},
  year          = {2023},
  eprint        = {2303.09295},
  archiveprefix = {arXiv},
  doi           = {10.48550/arXiv.2303.09295}
}

@misc{song2021ddim,
  author        = {Song, Jiaming and Meng, Chenlin and Ermon, Stefano},
  title         = {Denoising Diffusion Implicit Models},
  year          = {2021},
  eprint        = {2010.02502},
  archiveprefix = {arXiv},
  doi           = {10.48550/arXiv.2010.02502},
  url           = {https://openreview.net/forum?id=St1giarCHLP}
}

@misc{wu2025omnidfa,
  author        = {Wu, Shiyu and Li, Shuyan and Li, Jing and Liu, Jing and Wang, Yequan},
  title         = {Few-Shot Synthetic Image Attribution: Identifying Unseen Generators with Limited Samples},
  year          = {2025},
  eprint        = {2509.25682},
  archiveprefix = {arXiv},
  doi           = {10.48550/arXiv.2509.25682}
}

@misc{xu2025manipshield,
  author        = {Xu, Zitong and Duan, Huiyu and Wang, Xiaoyu and Cai, Zhaolin and Zhang, Kaiwei and Hu, Qiang and Liu, Jing and Min, Xiongkuo and Zhai, Guangtao},
  title         = {{ManipShield}: A Unified Framework for Image Manipulation Detection, Localization and Explanation},
  year          = {2025},
  eprint        = {2511.14259},
  archiveprefix = {arXiv},
  doi           = {10.48550/arXiv.2511.14259}
}

@article{yan2024jrc,
  author  = {Yan, Bosheng and Li, Chang-Tsun and Lu, Xuequan},
  title   = {{JRC}: Deepfake detection via joint reconstruction and classification},
  journal = {Neurocomputing},
  volume  = {598},
  pages   = {127862},
  year    = {2024},
  doi     = {10.1016/j.neucom.2024.127862}
}

@misc{yan2025sanity,
  author        = {Yan, Shilin and Li, Ouxiang and Cai, Jiayin and Hao, Yanbin and Jiang, Xiaolong and Hu, Yao and Xie, Weidi},
  title         = {A Sanity Check for {AI}-generated Image Detection},
  year          = {2025},
  eprint        = {2406.19435},
  archiveprefix = {arXiv},
  doi           = {10.48550/arXiv.2406.19435}
}

@misc{yang2025all,
  author        = {Yang, Zheng and Chen, Ruoxin and Yan, Zhiyuan and Zhang, Ke-Yue and Fu, Xinghe and Wu, Shuang and Shu, Xiujun and Yao, Taiping and Ding, Shouhong and Qin, Zequn and Li, Xi},
  title         = {All Patches Matter, More Patches Better: Enhance {AI}-Generated Image Detection via Panoptic Patch Learning},
  year          = {2025},
  eprint        = {2504.01396},
  archiveprefix = {arXiv},
  doi           = {10.48550/arXiv.2504.01396}
}

@misc{yang2025d3,
  author        = {Yang, Yongqi and Qian, Zhihao and Zhu, Ye and Russakovsky, Olga and Wu, Yu},
  title         = {{D$^3$}: Scaling Up Deepfake Detection by Learning from Discrepancy},
  year          = {2025},
  eprint        = {2404.04584},
  archiveprefix = {arXiv},
  doi           = {10.48550/arXiv.2404.04584}
}

@misc{zhou2025where,
  author        = {Zhou, Yuchen and Tang, Jiayu and Xiao, Xiaoyan and Lin, Yueyao and Liu, Linkai and Guo, Zipeng and Fei, Hao and Xia, Xiaobo and Gou, Chao},
  title         = {Where, What, Why: Towards Explainable Driver Attention Prediction},
  year          = {2025},
  eprint        = {2506.23088},
  archiveprefix = {arXiv},
  doi           = {10.48550/arXiv.2506.23088}
}

@inproceedings{zhou2018richfeatures,
  author    = {Zhou, Peng and Han, Xintong and Morariu, Vlad I. and Davis, Larry S.},
  title     = {Learning Rich Features for Image Manipulation Detection},
  booktitle = {Proceedings of the IEEE/CVF Conference on Computer Vision and Pattern Recognition (CVPR)},
  pages     = {1053--1061},
  year      = {2018},
  publisher = {IEEE},
  address   = {Salt Lake City, UT, USA}
}

@inproceedings{rossler2019faceforensicspp,
  author    = {R{\"o}ssler, Andreas and Cozzolino, Davide and Verdoliva, Luisa and Riess, Christian and Thies, Justus and Nie{\ss}ner, Matthias},
  title     = {FaceForensics++: Learning to Detect Manipulated Facial Images},
  booktitle = {Proceedings of the IEEE/CVF International Conference on Computer Vision (ICCV)},
  pages     = {1--11},
  year      = {2019},
  publisher = {IEEE},
  address   = {Seoul, Korea}
}

@inproceedings{li2020facexray,
  author    = {Li, Lingzhi and Bao, Jianmin and Zhang, Ting and Yang, Hao and Chen, Dong and Wen, Fang and Guo, Baining},
  title     = {Face X-Ray for More General Face Forgery Detection},
  booktitle = {Proceedings of the IEEE/CVF Conference on Computer Vision and Pattern Recognition (CVPR)},
  pages     = {5001--5010},
  year      = {2020},
  publisher = {IEEE},
  address   = {Seattle, WA, USA}
}

@inproceedings{wang2020cnnspot,
  author    = {Wang, Sheng-Yu and Wang, Oliver and Zhang, Richard and Owens, Andrew and Efros, Alexei A.},
  title     = {{CNN}-Generated Images Are Surprisingly Easy to Spot...for Now},
  booktitle = {Proceedings of the IEEE/CVF Conference on Computer Vision and Pattern Recognition (CVPR)},
  pages     = {8695--8704},
  year      = {2020},
  publisher = {IEEE},
  address   = {Seattle, WA, USA}
}

@inproceedings{qian2020f3net,
  author    = {Qian, Yuyang and Yin, Guojun and Sheng, Lu and Chen, Zixuan and Shao, Jing},
  title     = {Thinking in Frequency: Face Forgery Detection by Mining Frequency-Aware Clues},
  booktitle = {Computer Vision -- ECCV 2020},
  pages     = {86--103},
  year      = {2020},
  publisher = {Springer},
  address   = {Glasgow, UK}
}

@inproceedings{luo2021highfrequency,
  author    = {Luo, Yuchen and Zhang, Yong and Yan, Junchi and Liu, Wei},
  title     = {Generalizing Face Forgery Detection with High-Frequency Features},
  booktitle = {Proceedings of the IEEE/CVF Conference on Computer Vision and Pattern Recognition (CVPR)},
  pages     = {16317--16326},
  year      = {2021},
  publisher = {IEEE},
  address   = {Nashville, TN, USA}
}

@inproceedings{ojha2023univfd,
  author    = {Ojha, Utkarsh and Li, Yuheng and Lee, Yong Jae},
  title     = {Towards Universal Fake Image Detectors that Generalize Across Generative Models},
  booktitle = {Proceedings of the IEEE/CVF Conference on Computer Vision and Pattern Recognition (CVPR)},
  pages     = {24480--24489},
  year      = {2023},
  publisher = {IEEE},
  address   = {Vancouver, BC, Canada}
}

@inproceedings{liu2024fatformer,
  author    = {Liu, Huan and Tan, Zichang and Tan, Chuangchuang and Wei, Yunchao and Wang, Jingdong and Zhao, Yao},
  title     = {Forgery-aware Adaptive Transformer for Generalizable Synthetic Image Detection},
  booktitle = {Proceedings of the IEEE/CVF Conference on Computer Vision and Pattern Recognition (CVPR)},
  pages     = {10770--10780},
  year      = {2024},
  publisher = {IEEE},
  address   = {Seattle, WA, USA}
}

@misc{zhu2023minigpt4,
  author        = {Zhu, Deyao and Chen, Jun and Shen, Xiaoqian and Li, Xiang and Elhoseiny, Mohamed},
  title         = {{MiniGPT-4}: Enhancing Vision-Language Understanding with Advanced Large Language Models},
  year          = {2023},
  eprint        = {2304.10592},
  archiveprefix = {arXiv},
  doi           = {10.48550/arXiv.2304.10592}
}

@misc{liu2023llava,
  author        = {Liu, Haotian and Li, Chunyuan and Li, Yuheng and Lee, Yong Jae},
  title         = {Improved Baselines with Visual Instruction Tuning},
  year          = {2023},
  eprint        = {2310.03744},
  archiveprefix = {arXiv},
  doi           = {10.48550/arXiv.2310.03744}
}

@inproceedings{sun2025visuallinguistic,
  author    = {Sun, Ke and Chen, Shen and Yao, Taiping and Zhou, Ziyin and Ji, Jiayi and Sun, Xiaoshuai and Lin, Chia-Wen and Ji, Rongrong},
  title     = {Towards General Visual-Linguistic Face Forgery Detection},
  booktitle = {Proceedings of the IEEE/CVF Conference on Computer Vision and Pattern Recognition (CVPR)},
  pages     = {19576--19586},
  year      = {2025},
  publisher = {IEEE},
  address   = {Nashville, TN, USA}
}

@misc{radford2021clip,
  author        = {Radford, Alec and Kim, Jong Wook and Hallacy, Chris and Ramesh, Aditya and Goh, Gabriel and Agarwal, Sandhini and Sastry, Girish and Askell, Amanda and Mishkin, Pamela and Clark, Jack and Krueger, Gretchen and Sutskever, Ilya},
  title         = {Learning Transferable Visual Models from Natural Language Supervision},
  year          = {2021},
  eprint        = {2103.00020},
  archiveprefix = {arXiv},
  doi           = {10.48550/arXiv.2103.00020}
}

@misc{touvron2023llama2,
  author        = {Touvron, Hugo and Martin, Louis and Stone, Kevin and Albert, Peter and Almahairi, Amjad and Babaei, Yasmine and Bashlykov, Nikolay and Batra, Soumya and Bhargava, Prajjwal and Bhosale, Shruti and others},
  title         = {Llama 2: Open Foundation and Fine-Tuned Chat Models},
  year          = {2023},
  eprint        = {2307.09288},
  archiveprefix = {arXiv},
  doi           = {10.48550/arXiv.2307.09288}
}

@misc{hu2021lora,
  author        = {Hu, Edward J. and Shen, Yelong and Wallis, Phillip and Allen-Zhu, Zeyuan and Li, Yuanzhi and Wang, Shean and Wang, Lu and Chen, Weizhu},
  title         = {{LoRA}: Low-Rank Adaptation of Large Language Models},
  year          = {2021},
  eprint        = {2106.09685},
  archiveprefix = {arXiv},
  doi           = {10.48550/arXiv.2106.09685}
}

@inproceedings{kirillov2023sam,
  author    = {Kirillov, Alexander and Mintun, Eric and Ravi, Nikhila and Mao, Hanzi and Rolland, Chloe and Gustafson, Laura and Xiao, Tete and Whitehead, Spencer and Berg, Alexander C. and Lo, Wan-Yen and Dollar, Piotr and Girshick, Ross},
  title     = {Segment Anything},
  booktitle = {Proceedings of the IEEE/CVF International Conference on Computer Vision (ICCV)},
  pages     = {4015--4026},
  year      = {2023},
  publisher = {IEEE},
  address   = {Paris, France}
}

@misc{shao2024deepseekmath,
  author        = {Shao, Zhihong and Wang, Peiyi and Zhu, Qihao and Xu, Runxin and Song, Junxiao and Bi, Xiao and Zhang, Haowei and Zhang, Mingchuan and Li, Y. K. and Wu, Y. and Guo, Daya and others},
  title         = {{DeepSeekMath}: Pushing the Limits of Mathematical Reasoning in Open Language Models},
  year          = {2024},
  eprint        = {2402.03300},
  archiveprefix = {arXiv},
  doi           = {10.48550/arXiv.2402.03300}
}

@inproceedings{tan2024upsampling,
  author    = {Tan, Chuangchuang and Zhao, Yao and Wei, Shikui and Gu, Guanghua and Liu, Ping and Wei, Yunchao},
  title     = {Rethinking the Up-Sampling Operations in {CNN}-Based Generative Network for Generalizable Deepfake Detection},
  booktitle = {Proceedings of the IEEE/CVF Conference on Computer Vision and Pattern Recognition (CVPR)},
  pages     = {28130--28139},
  year      = {2024},
  publisher = {IEEE},
  address   = {Seattle, WA, USA}
}

@inproceedings{guo2023hifinet,
  author    = {Guo, Xiao and Liu, Xiaohong and Ren, Zhiyuan and Grosz, Steven and Masi, Iacopo and Liu, Xiaoming},
  title     = {Hierarchical Fine-Grained Image Forgery Detection and Localization},
  booktitle = {Proceedings of the IEEE/CVF Conference on Computer Vision and Pattern Recognition (CVPR)},
  pages     = {3155--3165},
  year      = {2023},
  publisher = {IEEE},
  address   = {Vancouver, BC, Canada}
}

@inproceedings{guillaro2023trufor,
  author    = {Guillaro, Fabrizio and Cozzolino, Davide and Sud, Avneesh and Dufour, Nicholas and Verdoliva, Luisa},
  title     = {{TruFor}: Leveraging All-Round Clues for Trustworthy Image Forgery Detection and Localization},
  booktitle = {Proceedings of the IEEE/CVF Conference on Computer Vision and Pattern Recognition (CVPR)},
  pages     = {20606--20615},
  year      = {2023},
  publisher = {IEEE},
  address   = {Vancouver, BC, Canada}
}

@inproceedings{zhang2023pal4vst,
  author    = {Zhang, Lingzhi and Xu, Zhengjie and Barnes, Connelly and Zhou, Yuqian and Liu, Qing and Zhang, He and Amirghodsi, Sohrab and Lin, Zhe and Shechtman, Eli and Shi, Jianbo},
  title     = {Perceptual Artifacts Localization for Image Synthesis Tasks},
  booktitle = {Proceedings of the IEEE/CVF International Conference on Computer Vision (ICCV)},
  pages     = {7579--7590},
  year      = {2023},
  publisher = {IEEE},
  address   = {Paris, France}
}

@inproceedings{lai2024lisa,
  author    = {Lai, Xin and Tian, Zhuotao and Chen, Yukang and Li, Yanwei and Yuan, Yuhui and Liu, Shu and Jia, Jiaya},
  title     = {{LISA}: Reasoning Segmentation via Large Language Model},
  booktitle = {Proceedings of the IEEE/CVF Conference on Computer Vision and Pattern Recognition (CVPR)},
  pages     = {9579--9589},
  year      = {2024},
  publisher = {IEEE},
  address   = {Seattle, WA, USA}
}

@inproceedings{chen2024internvl,
  author    = {Chen, Zhe and Wu, Jiannan and Wang, Wenhai and Su, Weijie and Chen, Guo and Xing, Sen and Zhong, Muyan and Zhang, Qinglong and Zhu, Xizhou and Lu, Lewei and others},
  title     = {{InternVL}: Scaling up Vision Foundation Models and Aligning for Generic Visual-Linguistic Tasks},
  booktitle = {Proceedings of the IEEE/CVF Conference on Computer Vision and Pattern Recognition (CVPR)},
  pages     = {24185--24198},
  year      = {2024},
  publisher = {IEEE},
  address   = {Seattle, WA, USA}
}

@misc{wang2024qwen2vl,
  author        = {Wang, Peng and Bai, Shuai and Tan, Sinan and Wang, Shijie and Fan, Zhihao and Bai, Jinze and Chen, Keqin and Liu, Xuejing and Wang, Jialin and Ge, Wenbin and others},
  title         = {{Qwen2-VL}: Enhancing Vision-Language Model's Perception of the World at Any Resolution},
  year          = {2024},
  eprint        = {2409.12191},
  archiveprefix = {arXiv},
  doi           = {10.48550/arXiv.2409.12191}
}

@misc{wang2025opensdi,
  author        = {Wang, Yabin and Huang, Zhiwu and Hong, Xiaopeng},
  title         = {{OpenSDI}: Spotting Diffusion-Generated Images in the Open World},
  year          = {2025},
  eprint        = {2503.19653},
  archiveprefix = {arXiv},
  doi           = {10.48550/arXiv.2503.19653}
}

@misc{nataraj2019cooccurrence,
  author        = {Nataraj, Lakshmanan and Mohammed, Tajuddin Manhar and Chandrasekaran, Shivkumar and Flenner, Arjuna and Bappy, Jawadul H. and Roy-Chowdhury, Amit K. and Manjunath, B. S.},
  title         = {Detecting {GAN} Generated Fake Images Using Co-Occurrence Matrices},
  year          = {2019},
  eprint        = {1903.06836},
  archiveprefix = {arXiv},
  doi           = {10.48550/arXiv.1903.06836}
}

@inproceedings{zhang2019freqspec,
  author    = {Zhang, Xu and Karaman, Svebor and Chang, Shih-Fu},
  title     = {Detecting and Simulating Artifacts in {GAN} Fake Images},
  booktitle = {2019 IEEE International Workshop on Information Forensics and Security (WIFS)},
  pages     = {1--6},
  year      = {2019},
  publisher = {IEEE},
  address   = {Delft, Netherlands}
}

@misc{chai2020patchfor,
  author        = {Chai, Lucy and Bau, David and Lim, Ser-Nam and Isola, Phillip},
  title         = {What Makes Fake Images Detectable? Understanding Properties that Generalize},
  year          = {2020},
  eprint        = {2008.10588},
  archiveprefix = {arXiv},
  doi           = {10.48550/arXiv.2008.10588}
}

@inproceedings{tan2023ldgard,
  author    = {Tan, Chuangchuang and Zhao, Yao and Wei, Shikui and Gu, Guanghua and Wei, Yunchao},
  title     = {Learning on Gradients: Generalized Artifacts Representation for {GAN}-Generated Images Detection},
  booktitle = {Proceedings of the IEEE/CVF Conference on Computer Vision and Pattern Recognition (CVPR)},
  pages     = {12105--12114},
  year      = {2023},
  publisher = {IEEE},
  address   = {Vancouver, BC, Canada}
}

@article{tan2024freqnet,
  author    = {Tan, Chuangchuang and Zhao, Yao and Wei, Shikui and Gu, Guanghua and Liu, Ping and Wei, Yunchao},
  title     = {Frequency-Aware Deepfake Detection: Improving Generalizability through Frequency Space Domain Learning},
  journal   = {Proceedings of the AAAI Conference on Artificial Intelligence},
  volume    = {38},
  number    = {5},
  pages     = {5052--5060},
  year      = {2024},
  doi       = {10.1609/aaai.v38i5.28310}
}

% \clearpage
% \appendix
% \input{supplementary/arxiv_appendix}

\end{document}